# A New Approach for Optimizing Highly Nonlinear Problems Based on the Observer Effect Concept

Mojtaba Moattari[1], Emad Roshandel[2], Shima Kamyab[3], Zohreh Azimifar[4]

[1] Department of Electrical and Computer Engineering, Amirkabir University of Technology, Tehran, Iran, moatary@aut.ac.ir
[2] Department of Research and Development, Isfahan University of Technology, Shiraz, Iran, e.roshandel@esmg.co.ir
[3] Department of Computer Engineering, Shiraz University, Shiraz, Iran, shima.kamyab@gmail.com
[4] Department of Computer Engineering, Shiraz University, Shiraz, Iran, azimifar@cse.shirazu.ac.ir

**Abstract**- A lot of real-world engineering problems represent dynamicity with nests of nonlinearities due to highly complex network of exponential functions or large number of differential equations interacting together. Such search spaces are provided with multiple convex regions peaked with diverse nonlinear slopes and in non-homogenous ways. To find global optima, a new meta-heuristic algorithm is proposed based on Observer Effect concepts for controlling memory usage per localities without pursuing Tabu-like cut-off approaches. Observer effect in physics (/psychology) regards bias in measurement (/perception) due to the interference of instrument (/knowledge). Performance analysis of the proposed algorithms is sought in two real-world engineering applications, i.e., Electroencephalogram feature learning and Distributed Generator parameter tuning, each of which having nonlinearity and complex multi-modal peaks distributions as their characteristics. In addition, the effect of version improvement has been assessed. The performance comparison with other optimizers in the same context suggests that proposed algorithm is useful both solely and in hybrid Gradient Descent settings where problem's search space is nonhomogeneous in terms of local peaks density.

## 1. Introduction:

Real-world engineering problems often contain search space with localities having different densities of peaks. Lower homogeneity of a locality in term of peak density makes it harder to track lower-cost-solutions from gathered knowledge about cost points. Therefore, optimizers that keep track of the extent of homogeneity is essential for such problems.

Machine Learning models may comprise of many nested exponentials because of the network of interacting activation functions. Engineering systems like Power Stabilizers may contain nonlinear differential equations entangled together, causing unexpected nonlinearity and dynamicity salient in the resulting search space of observable system function. Therefore, having a Gradient Descent (GD) initializer or a metaheuristic parameter tuner that searches more randomly (less knowledge-driven) in search localities with more fluctuated fitness, is of interest. Hence, a mechanism for localized control of memory usage is sought in this work.

To find the metaheuristics that adapt memory usage degree locally, some related works are discussed here. Tabu Search methods (TS) are specially designed to reduce traverse of infertile search regions in optimization problems [1]. Use of regions as taboo causes the method to reduce the chance of revisiting regions. As a result, other promising regions will have higher chance to be traversed. Up to now, many algorithms, being powerful at exploration, are combined with TS to have better exploitation capability. Hybrid Nelder-Mead and Tabu

Search, simulated annealing [2] and Tabu Search [3], hybrid ant colony optimization and Tabu Search [4] attracted more attentions among these algorithms. However, Tabu search is not locally adaptive, i.e. retaining behavior even after seeking new observations. Moreover, its computational burden is too high for large scale engineering problems. To tackle this issue, various memory-based optimization techniques [4] are developed to seek new solutions based on current context [50, 51]. Some methods like Estimation of Distributions Algorithm (EDAs) introduced by Mühlenbein and Paaß, intend to estimate the probability distribution of solutions in a locality and generate solutions by sampling the PDF. The major concern of these algorithms lies in memory inductions, i.e. model fitting using solutions in memory or pdf estimation [5-10]. Not only have they large space complexity in large-scale algorithms, but also they need a lot of computations. Yet, there is no method to fully address issues of where, when, and how much memory should be used. These questions are crucial in the context of engineering search spaces having dynamicity and nonlinearity from one locality region to the other one.

Ant colony optimization (ACO) is another approach that uses memory elements in its solution updating process [11-13]. It is inspired from the foraging behavior of ants. For better searching for food, ants secrete a substance called pheromone to signal path of food to each other. Pheromone will remain on the ground only for a definite time. That said, the ACO algorithm can act in a more systematic way when there is pheromone. Otherwise, it performs random searches. In ACO, pheromone evaporation rate can control the extent of memory usage. However, ACO algorithms have no adaptive

Authentic studies have not yet proposed a more adaptive framework focusing solely on a cluster-based limitation of memory-usage. Tabu Search methods cut off the possibility to search near solutions that are in its tabu set, while our method decreases the likelihood of solution update in a region but doesn't make it zero. Moreover, Tabu Search neither adapts clusters in choosing method-based search versus random searching, nor controls locality volume. To the best of authors' knowledge, no algorithms yet have been proposed to dynamically control the useable memory extent with different parameter settings per different localities.

Therefore, a new algorithm, inspired by the observer effect, is proposed to impose taboo degree of localities in a soft way, automatically adapt innovation rate in exploration, and also choose a locally adaptive memory filtering approach. The observer effect is mainly discussed in physical systems, instrumentation and social psychology. More or less, this concept is considered in other branches like politics, economy, and sociology. Generally, the observer effect is associated with changes imposed on the observable entity through observation. This effect can also lead to the wrong measurement of entity properties. For the case of a physical process, active measurements can affect the system which is being measured. For the example of electron detection, a photon has to interact with it [49]. This active measurement distorts the electron's properties and eventually leads to a distorted measurement. For more detailed information on Observer Effect in physics, the reader can refer to [53, 48 and 49]. By looking at optimization problems as measurement problems, the effect of observer can be defined and controlled. By controlling the extent of memory usage on predicting more fit solutions, the extent of observer effect is controlled adaptively for the problem. For better tackling locations in search space that past knowledge of solution distribution is not helpful and informative for finding better solution, Observer Effect Optimization (OEO) is proposed. Its purpose is to manage where to use past knowledge and where not to. In the following sub-section, further explanations about various scientific origins of the proposed algorithm have been brought about.

To tackle the aforementioned engineering problems, it is necessary to control the amount of knowledge usage per locality and exploit less in localities that are less certain and homogeneous than it has been already supposed. When the exploitation process per locality fails frequently at finding a lower cost during solution update, our proposed algorithm selects solutions from other localities. Knowledge transfer rate control during solution update process is the main idea of proposed OEO method.

Speaking psychologically, observer (expectancy) originates out of bias in knowing a case of study [17, 22, 23]. Previous knowledge about a process can lead to a hindrance about further knowing of the process [18]. A clear example of this bias [19] is evident in backmasking of pieces of music. Music is backmasked when its track is played backwards. The reversely played music is usually meaningless and composed of random sounds. However, having a biased interpretation in some specific utterance, one may mistakenly think it entails meaningful hidden messages. To put the psychological view of observer effect into the mentioned general definition, one can regard observed system as one's realization about a subject, assuming observer as ability to realize and accumulate knowledge. When the knowledge interferes, the realization will be distorted. However, in cases of sufficient knowledge, presuming that as the truth of observable process will not cause misinformation and saves time and effort [21]. As the Brain manages the extent of the observer effect in observations, it is possible to bring this management into the domain of meta-heuristic optimization. The optimizer might use the knowledge to save energy in cases knowledge is enough, or search randomly if the knowledge is not helpful.

Two real-world problems are used to evaluate the performance of OEO. The first problem is parameter initialization of GD optimizer which is a subspace learning method for feature extraction of Electro-Encephalogram (EEG) speech imagery data. The second problem tunes parameters of Distributed Generators (DGs) as their sizes and locations which is a complex objective function among other Power System problems due to lack of analytical solutions. The commonality of both case studies is their specific optimization search spaces that are not only multimodal but also hard to track localized behavior of function. Such problems can apply OEO to dynamically switch between uniform-random-search and elite-nearby-search by modifying localities' hyper-parameter settings during runtime.

Accordingly, the objective of the proposed OEO algorithm is to tackle an optimization problem as a measurable system and view an optimization algorithm as an observer. As an observer must not distort the measurable system, the optimizer's accumulated knowledge of sought solutions, must not misdirect the optimization process. Eventually, OEO uses knowledge in a region as long as it leads to fitness improvement or otherwise ignores it.

The main contributions of this paper are:

- Introducing a new algorithm to control the memory effect in a soft way with no memory-usage cut-off. The algorithm is designed for real-world search spaces that have high dynamicity and nonlinearity. The search space in these problems comprise of multiple convex regions that are scattered in non-homogenous ways.
- Evaluating algorithm as parameter initializer of GD optimization of the nonlinear objective function over highly complex and dynamic data: a newly proposed objective function that does spatial filtering on EEG data and optimizes parameters in a hybrid mode with GD for better classification accuracy.
- Evaluating algorithm on a parameter tuning study with highly structured multi-modal solutions: selecting the location and size of the DGs in a power grid.

Section 2 explains the OEO method fundamentals and basis. Section 3 gives out a modified version for the algorithm that makes cluster parameters specialized. In Section 4, two case studies are introduced in detail and evaluation results are shown and analyzed at the end of each sub-section. Finally, the conclusion ends the paper by related challenges.

## 2. The proposed method, Observer Effect Optimization (OEO):

To be precise, the proposed OEO is a tool to control the effect of existing knowledge into the decision of new solutions. Here existing knowledge of solution distributions may be interpreted as observer information. Eventually, the OEO sets up a competition between a random selection of solutions and their memory-based selection with different tuned parameters for each region. In this paper, all memory-based solution selection methods use gathered knowledge of sought solutions.

The proposed algorithm uses priori information in regions which the result gets improved and uses random updating information in other regions. Realization of this approach is feasible through adaptive parameter discerning which is suitable among memory or random based solution production. This process is performed in version 1 of the proposed algorithm. In version 2, different adaptive parameters are selected for different regions.

Moreover, by producing or updating solutions in more promising regions, a Tabu like scent is given to the algorithm which automatically sets up a competition among clusters. Characters of the proposed OEO are listed as below:

- Discerns how much algorithm must trust past information, distribution of solutions, and previously sought costs.
- Helps to seek for random solutions and search in other regions of space as searching in previously sought clusters will not lead to better solutions.
- Leads to the interplay between random searching and model-based searching.
- Updates solution near elites using regular records of elites probability.
- Chooses between exploration and exploitation in an adaptive way by updating 3 observer parameters.

### 2.1. The OEO, General Framework

**Definition 1. OEO**
OEO is denoted by $O(T, 1, A, B, G, E, c_j, S)$, with T as iteration count, $\alpha$ representing the cluster selection probability, $B$ being as cluster selection probability, and $\gamma$ as random cluster update probability. In each iteration t<T, O generates and selects new cluster $c_j$ ( j>len(c) ) with probability $A$ or selects previous clusters with probability 1-$\alpha$. Afterwards, it selects one cluster based on its effectiveness $E_j$ (overall sought fitness). Random solution selection is conducted by probability $G_j$, or gets linear combination of best sought fitness solutions in the area with probability 1-$G_j$, $\beta$ is a global parameter as a nonlinear scaler of E, i.e. for controlling the diversification (ignorance of priori knowledge or observer effect). S is a sampler.

Using the insight gathered from scientific background of observer effect, the idea behind the proposed OEO algorithm is stated as follows. Considering the search space *S,* The OEO algorithm starts with an empty population of solutions. Every iteration of the OEO algorithm includes generating a new solution *ind_i*, randomly or using the notion of observer effect with a probability. The first iterations, accordingly, include randomly generating the individuals and account them as some *cluster centers*. If $c_i$ denotes the $i^{th}$ cluster center and $x_{mn}, x_{mx}$, stand for the maximum and minimum values allowed in the *d-dimensional* search space, The generation of new cluster will be defined as $c_i = x_{mn} + rand(1,d) * (x_{max} - x_{min})$.

OEO controls the effect of observer information into a decision of new solutions. Observer information is past knowledge and prejudice about a possible model of solutions distribution. This sets up a competition between the random selection of solutions and their memory-based selection. Algorithm 1 includes the main framework of proposed OEO in each iteration. *A*, *B*, and *G* parameters, adaptively control the extent of knowledge/memory-biases entrance in randomized exploration, localities-pruned exploration, and elites-induced exploitation respectively. They will be adapted during algorithm run for better performance of the run. It must be noted that the bigger the *A* and *B*, the lower the observer effect.

**Algorithm 1. The main framework of the OEO.**
**Input:**
    Number of clusters ($N_c$), A, G, B, Preferred cluster effectiveness metric, Preferred solution update metric
**Output:**
    Least cost solution
**Process:**
    While Termination Criteria not met, do:
        If rand>A,
            Randomly create new solution globally and add this as new cluster mean.
        Else,
            Select a cluster using a preferred metric with sensitivity B. (Refer to Section 2.3)
            If number of cluster solution<4 or rand< G,
                Add solution randomly to cluster
            Else,
                Add solution using a preferred method over the solutions in the cluster (ex. weighted mean of data)
        Check out the new solution's cost.
        Adapt *G*
        Adapt *A*
        Adapt *B*
        After a little while, (each 4 iteration), remove solutions with lowest cost
    Remove empty clusters.

After the adaptation of parameters, the remaining lowest cost solutions will be removed to retain a maximum number of solutions during each iteration.

Algorithm possesses 3 control parameters called *A, B*, and *G* which manage usage of memory and gathered knowledge on finding new solution location. *A* controls the rate of global searching versus local, in searching within clusters. B controls how much clusters scores should be considered when a Roulette Wheel Selector selects one cluster among others. Finally, *G* parameter controls the probability of rule-based update over the random selection of solution in a locality. All these three parameters are controlled adaptively as optimization goes on. Deeper explanations and parameter adaptation of *A, B* and *G* is available in sections 2.3, 2.2 and 2.4, respectively. More detailed comparison of solution update types of the proposed OEO is available in Figure1. In the figure, circles represent clusters which are brought by sphere cluster creator which is detailed in the next section.

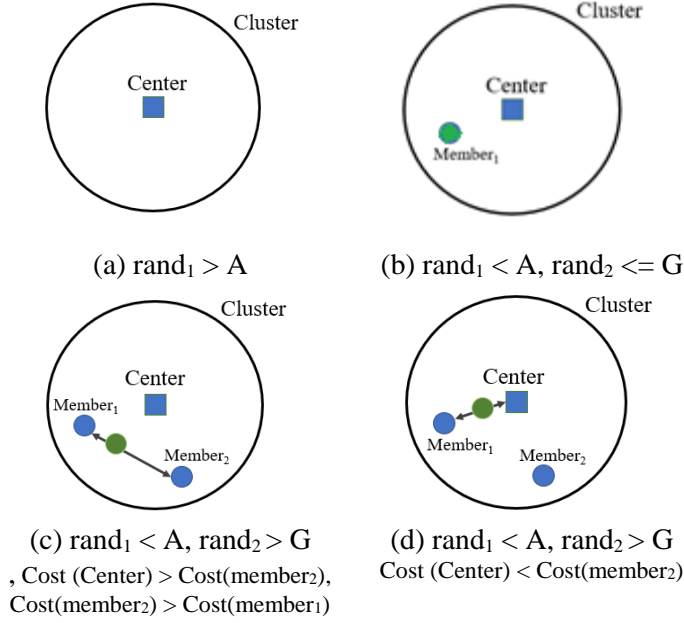

**Figure 1: Types of solutions in the proposed OEO method (algorithm 1). In all four cases, green circle is the new solution. a) If rand₁ is higher than A, a new cluster with its center as a new solution will be generated. b) In this case, a new solution will be generated randomly in an existing cluster. c, d) Solution is updated based on a metric that selects and combines existing elites in a cluster of interest.**

### 2.2. Controlling rate of global searching (*A*)

The outer decision layer of the OEO during each iteration deals with deciding whether or not to use memory. This decision takes place by a uniform random number generator. When the generated number is lower than *A* (0<*A*<1), the algorithm reduces search region to existing clusters; preventing algorithm to search globally. The higher the value of *A*, the higher the use of memory, the higher the concerning observer effect will also be.

Adaptation of *A* should be in such a way that leads to a more sensitive algorithm considering its current progress. Increasing and decreasing *A* during run time can fulfill that objective. The process of updating A is as below:

- Decrease A randomly if adding into cluster failed, otherwise decrease it based on a previous knowledge (e.x., elites nearby).
- Failure takes place when it does not lead to a better solution in the cluster.

### 2.3. Controlling the amount of nonlinearity for cluster selection (*B*)

After bypassing random searching phase, the algorithm has been entered to state which one of the clusters should be selected for local searching. This selection is implemented using a roulette wheel search (RWS) method. RWS selects an element based on its probability in the Probability Distribution. It is derived by normalizing cluster effectiveness obtained through available metrics in Section C. By passing the distribution through the following nonlinearity, the certainty of cluster selection will be affected. Therefore, the intensity of involvement of memory in area selection will be controlled. The nonlinearity function is shown in (1).

$$f(p(x), B) = e^{\left(\frac{\log(p(x))}{\log(\exp(1) + \lambda * B)}\right)} \quad (1)$$

Where p(x) is normalized effectiveness of each cluster and $\lambda$ is set to 18. *B* controls the probability of selection of a cluster among other clusters based on knowledge about its effectiveness. By increasing *B*, elements of *f(p(x), B)* get closer to each other; making the past knowledge effectless to region selection. Taking logarithm from both sides of the equation (1) makes its purpose clearer. In the logarithmic domain, as the function tries to impose a reduction to scale of log(p(x)) by function of B (with tunable sensitivity $\lambda$), the whole function undergoes an exponential scaling dependent on dynamic state of B in the runtime. This exponential rescaling gives B more ability to nonlinearly rescale p(x) compared to normal rescaling as it takes radical of probability items with root $\lambda B$ while retains p(x) monotonicity. Therefore, raising *B* leads to lower observer effect or more deviation between lowest and highest cluster effectiveness, resulting in more certain selection by RWS.

For instance, for cluster_effectiveness= [1 10 2 4 45 60 100 40 85 170 150 300 250], effect of nonlinearity function is represented in Fig. 2. It is evident that the lower the B, the higher the nonlinearity and so the higher the knowledge effect.

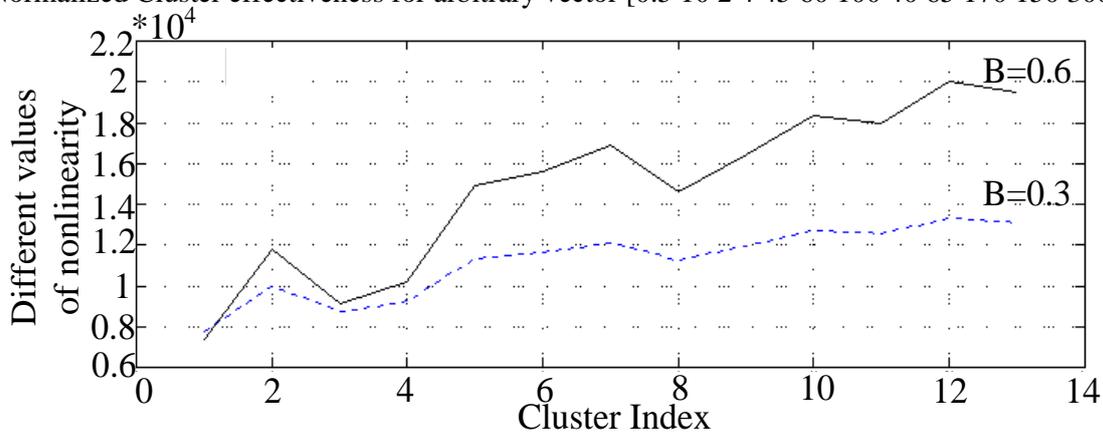

**Figure 2 - Effect of output effectiveness after getting passed from nonlinear strengthening function with two different *B* parameters. The higher the *B* value, the higher the ranges and eventually the higher the knowledge effect out of cluster effectiveness.**

*B* is adapted to adjust algorithm sensitivity in deciding when and where to use memory. By controlling *B*, the algorithm can automatically tune the rate of using knowledge for the best of itself. That is, after cluster selection and solution update, if it led to better fitness, Increase *B* by $m_1$, otherwise decrease it by $m_2$. $m_2$ is chosen less than $m_1$ due to the fact that better result is usually rare. This approach leads to an increase of *B* and thus more localization certainty as long as fitness increases and otherwise decreases *B* to stop further intensification when improper.

### 2.4. Controlling rate of rule-based update (*G*)

This parameter makes a choice between random solution selections in a cluster versus knowledge-based rule-based solution update. The metrics available for the rule-based update is put in section *C*. During each iteration, the decision is taken place using a parameter *G* and a generated random number. If the random number is higher than *G*, it randomly selects a solution in the concerning cluster region; otherwise, it performs method-based selection using a preferred metric among Table 1 methods. In Table 1, five rule-based solution update methods are proposed. They use currently sought solution of a region to produce a new solution.

Adaptation process for *G* is to decrease *G*, if method based update failed or random based addition succeeded. Else, increase. Success is leading to better fitness.

**Table 1- Cluster Solution Update Methods. In all cases, $X_{best}$ is the best solution in the intended hull. u is a uniform generated a random number between zero and one. $X_{start}$ is an arbitrary select solution in concerning hull. $X_1$ and $X_2$ are Randomly choose two solutions in concerning hull with fitnesses $f_1$, $f_2$, respectively. $X_{ibest}$ denotes the i$^{th}$ best solution in the hull.**

| Description | Method Tag | Description/ Pseudocode | Suitability |
|---|---|---|---|
| **The overall sum of all fitnesses in concerning cluster** | 'EitherRandomlyOrThroughBest' | If rand < p<br>    $X_{new}$ = Random solution<br>Else<br>    $X_{new}$= $X_{start}$ + u * ($X_{best}$- $X_{start}$) | • Combination of random and structured searching. More flexible if p is tuned during the run |
| **Select new solution near in sequel of best solution of the hull and other selected one.** | 'MoveThroughBest' | $X_{new}$= $X_{start}$ + u * ($X_{best}$- $X_{start}$) | • Best currently sought solution maybe closer to a best global solution. |
| **Select two solutions and choose a new solution in the sequel.** | 'Select2Sols&ChooseOneBetween' | $X_{new}$= $aX_1$+(1-a)$X_2$=$X_2$+($X_1$-$X_2$)*a | • Low fitness solutions get the filtered end of each iteration.<br>• By choosing a solution between two existing solutions, fitness improvement may be possible. |
| **Mean of solution locations in cluster.** | 'ClusterMean' | $X_{new}$ = mean of solutions contained in hull. | N/A |
| **mean of N best solutions** | 'MeanOfElites' | $X_{new}$ = $1/N \sum_{i=1}^{N} X_{ibest}$, $N = Constant$ | • Mean of elites' locations may give a suitable estimate of the better unsought solution. |
| **the weighted mean of solutions** | 'GetWeightedMeanOfSols' | $X_{new}$ = $f_1 X_1 + f_2 X_2$ | • New solution will be selected near better solutions of the hull.<br>• Differs from MeanOfElites case in the fact that non-elite solutions will affect the selection of location. |
| **N best solutions** | 'GetWeightedMeanOfElites' | $X_{new}$ = $\sum_{i=1}^{N} f_i X_{ibest}$, $N = Constant.$ | In comparison with MeanOfElites case, The better the solution, the closer the new solution gets to. |

Comparing adaptation trend of *G, B*, and *A* may clarify points of this approach. *G* sets up a competition among localized rule-based solution update versus localized random solution update. In the case of *B*, the competition is among different locations using the knowledge. *A* makes a competition between diversification and intensification with controlled memory usage. As they are three independent viewpoints of a problem, they may be suitable complementary.

Suitable cluster effectiveness metrics are proposed which are used in the algorithm are presented in Table. 2.

**Table 2- Cluster Effectiveness metrics**

| Description | Metric Tag in the algorithm | Suitability |
|---|---|---|
| **Overall sum of all fitnesses in concerning cluster** | 'SumFitnessPerVolume' | • High amount of fitness values in a region of search space, can be a sign of closeness of the region to the global solution. |
| **Sum of best solution fitness per volume** | 'SumEliteFitnessPerVolume' | • Instead of counting all fitnesses in computing effectiveness, only top solutions are selected. |
| **Best fitness value per volume** | 'BestFitnessPerVolume' | • Best solution fitness in each volume with any size can be a sign for the effectiveness of locating a global solution in that volume. |
| **Variance of fitness values per volume** | 'VarFitnessPerVolume' | • In a hull where spread of fitness values is high, it is more likely to find a better solution. |

For the modified version explained in the next section, due to locality dependency of algorithm hyperparameters, the algorithm may witness performance reduction for large scale parameter tuning. The reason is that total number of clusters is proportional to number of parameters which is logarithmic proportion to space scale. Such cluster number specification, may have negative impact on optimization effectiveness and deserves a numerical analysis in our future works.

## 3. Modified Observer Effect Optimization (M-OEO)

In the last section, the main framework of OEO is explained, adaptive parameters have been cleared out, and feasible metrics for memory-based solution update have been gathered as a table. In this section, two kinds of OEO versions are provided. They differ in selected metrics and individualizing gamma. After introducing these versions of the algorithm, a finalized flowchart of the modified version will be shown in Figure 3.

### 3.1. OEO, First version versus the modified version

By choosing suitable metrics for the generalized framework and using approaches in Sections B.2 as parameters adaptation method, the first version will be defined. Suitable metrics are chosen among metrics with the best average number of function evaluations among other metrics.

Algorithm metrics and parameter preferences are shown in Table 3. Cluster solution update metric and cluster effectiveness metric for each version of OEO is fixed and had undergone no tuning processes. Hyper-parameters $m_1$ and $m_2$ are tuned from set $\{0.1*n | 20 < n \in N <= 50\}$. The tuning process has selected hyper-parameter with the least average cost over 20 different runs per each optimizer version. In the first version, there is only one G parameter for all clusters. In the second version, each cluster has their own G parameter. This helps choosing between memory-based local searching over random local search. To initialize basic OEO hyperparameters (A, B, and G), the values are randomized for 50 different experiments for OEO and M-OEO and no significant difference between their initialized states are witnessed as the Odds ratio for best cost distribution w.r.t each hyperparameters was close to 1 (over 0.95). Therefore the values are selected randomly and similarly in both algorithm versions.

**Table 3- Algorithm metrics and parameter preferences. Most suitable parameters for algorithm are selected and shown.**

| Proposed algorithm | Associated metrics | | | Tuned parameter preferences | | G parameter | Initialized Num. Of Clusters | Basic OEO hyper-parameters' initial state ($1_Q$ is Q-tuple row vector with all values of 1.) |
|---|---|---|---|---|---|---|---|---|
| | Solution selection Metric | Cluster Solution Update metric | Cluster Effectiveness metrics | $m_1$ | $m_2$ | | | |
| SIMPLE OEO[ST] (OEO) | RWS | 'MeanOfElites' | 'MeanElite Fitness' | 5.4 | 3.9 | Scaler, Unique for all clusters | $N_c$= 10 | $A_{start}$=0.3 $B_{start}$=0.16 $G_{start}$=0.2 |
| M-OEO[ND] (Modified-OEO) | RWS | 'GetWeighted MeanOfSols' | 'SubtractFrom NearestFitness' | 7.4 | 5.6 | Vector with length of clusters count, Individualized for each cluster | $N_c$= 10 | $A_{start}$=0.3 $B_{start}$=0.16 $G_{start}$=0.2*$1_{Nc}$ |

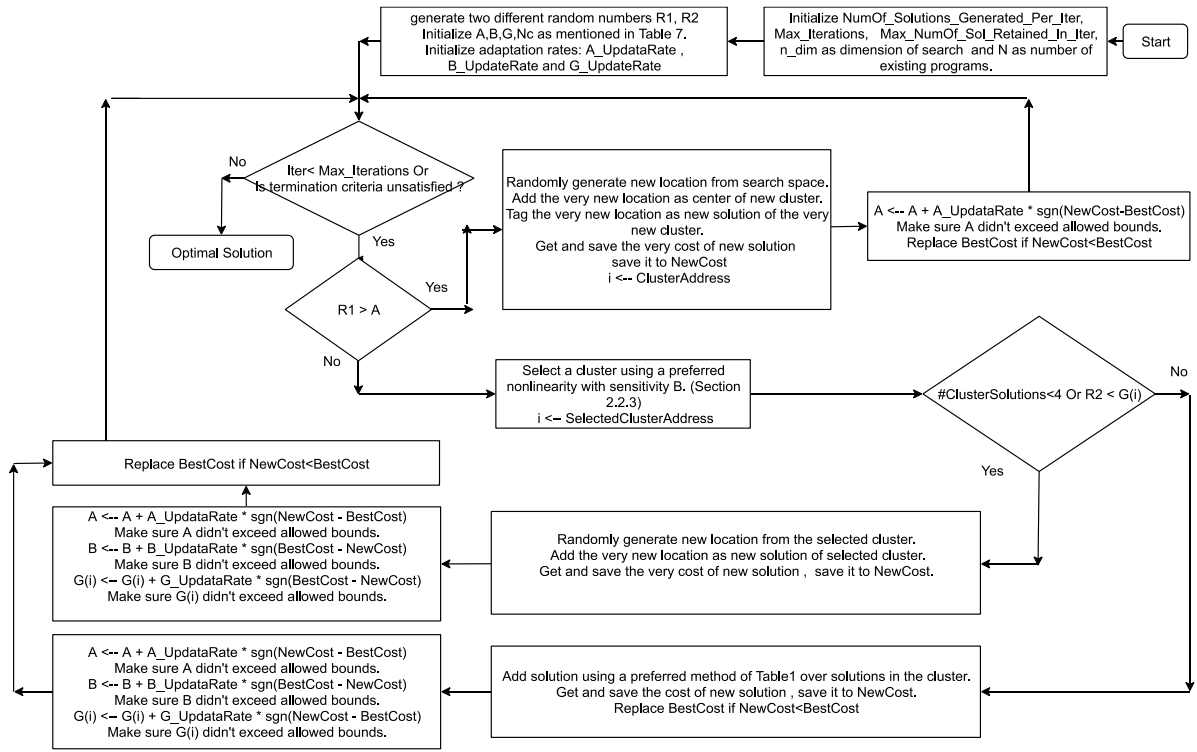

**Figure 3. Proposed flowchart of the modified version of OEO (M-OEO)**

## 4. Evaluation and analysis of two real-world problems

In this section, the OEO is evaluated on two real-world scenarios; one assesses the OEO capability in a machine learning study and the other in a system engineering study:

- The machine learning study is a subspace-filtering problem for optimizing the classification accuracy of brain-computer-interfaces. It demands a global optimization in its nonlinear objective function with multiple convex localities. Hence it needs a hybrid optimizer with GD as its main algorithm and our proposed OEO as its random initializer preventing the optimizer from trapping into local minima. This case is discussed in Section 4.1.

- On the other hand, the respective system engineering study is about a power system analyzer that aims to distribute optimal generators specifications using an objective function that keeps the system economically justified. So this is a parameter tuning scheme and should be able to search for better parameters in objectives that neither have a gradient for fixpoint solving purposes nor have a closed-form solution. Therefore, generality of the problem is NP-hard and the objective is to assess OEO capability on the study over other simplistic and commonly-used optimizers. This case is discussed in Section 4.2.

### 4.1. Parameter Learning Study, Nonlinear feature learning

To evaluate our proposed OEO algorithm, a nonlinear objective function based on Spatial Filtering is proposed. Spatial Filter in this study is Common Spatial Pattern [37, 39] mainly used for feature selection in EEG classification problems. Our objective function has terms w.r.t. which the function is nonconvex. As a result, GD cannot asymptotically converge to the global solution.

#### 4.1.1. The problem statement

The problem is to optimize the subspace filtering objective function using hybrid of OEO and GD methods, in which OEO is used as a GD initializer (OEO-GD), and to compare with other hybrid methods of GD. The GD method has been initialized by OEO. It has been evaluated on a feature filtering problem which is a weighted sum of each trial's Common Spatial Pattern (CSP). So, the concept of CSP will be described in the next section and afterwards, the main objective function will be derived.

#### 4.1.1.1. Common spatial pattern (CSP)

CSP is a feature extraction method for classifying EEG states in Brain Computer Interface. Due to its success in many neuroscientific case studies, it has turned into one of the most commonly used approaches for EEG features dimensionality reduction. It mainly seeks for a sub-space with maximum variance in one class and minimum variance in other class labels [37, 32]. In an evoke-related potential (ERP) with N as the channels count and T as the samples count, suppose that E is an $N \times T$ matrix of zero-mean data. The covariance matrix per class is derived by taking the mean of covariance matrices corresponding to that class over all trials. CSP finds set of orthogonal bases $w_k$ for $0 < k \in N < K_{components}$ such that:

$$w_k = \text{ArgMax}_{w\_k} \frac{w_k' C_i w_k}{w_k' (C_1 + C_2) w_k} \tag{2}$$

Where each $w_k$ is normalized automatically due to the formulation of the problem. i is class label index which is 1 or 2 in this case. C is different for each class and is derived by $E^T E$.

Equation (2) tries to find a representation which maximizes covariance of one class while minimzing the covariance of the other class. It is similar to the Linear Discriminant Analysis objective function in term of formulation and optimization approach. For comprehensive details on biological underpinnings of Eq. (2), the reader can refer to [37]. This equation turns into a generalized eigenvalue problem by setting the denominator as Lagrange multiplier to the numerator [39]. Taking derivative w.r.t. $w_k$ results in :

$$W \Lambda W' = (C_1 + C_2)^{-1} C_i (C_1 + C_2) \tag{3}$$

Where W is matrix out of $w_k$ as its columns and $\Lambda$ is a diagonal matrix of Lagrange multipliers in (3). Eq. (3) can be solved by power iterations or other SVD-based methods.

To show the performance of OEO-GD, some of the existing works regarding the subject of trials-filter-learning have been chosen as comparison baselines. The filter is CSP in this context. The better weighting of each EEG trial improves the filtering accuracy and as a result, improves the generalization capability of filtered subspace in classification. Therefore, a new filtering objective function is proposed and it is optimized with OEO-GD.

### 4.1.1.2. Devlaminck's work (CCSP)

The method which is suggested by Devlaminck et al., regards a correspondence between the spatial filters extracted from different subjects and aims to extract such correspondence with the help of a shared global basis $w_0$ and save the remnant in the subject-specific part $v_i$ [41].

$$w_i = w_0 + v_i, \tag{4}$$

He designed a particular objective function to learn the parameters

$$\sum_{k=1}^{K} \sum_{s=1}^{S} a_{sk} \frac{w_{sk}^T \Sigma_s^{(1)} w_{sk}}{s_{sk}^T \Sigma_s^{(2)} w_{sk} + \lambda_1 \|w_k\|^2 + \lambda_2 \|v_{sk}\|^2}, \quad a_{sk} \epsilon \{0,1\}, \quad \sum_{k=1}^{K} a_{sk} = 1 \tag{5}$$

Due to binary weights in (5), the weights cannot be approximated by GD. The idea of disintegrating the filter to subject-specific and subject-independent segments makes this method a useful comparison baseline for the cost function proposed in 5.1.1.4.

### 4.1.1.3. Regularized Common Spatial Pattern (RCSP)

This technique, Lotte's work, accounts for the covariance matrices of all subjects for a specific one [42]. The purpose is to use the shared information between subjects. The covariance matrix $\Sigma_{i*c}$ for i'th subject is as follows:

$$\widetilde{\Sigma}_{i^*,c} = (1 - \lambda)\Sigma_{i^*,c} + \frac{\lambda}{n-1}\sum_{i=1}^{n-1} \Sigma_{i^*,c} \tag{6}$$

Where c is the class per subject, and $\lambda \in [0,1]$ is a regularization parameter to control the shared covariance effect. Due to informative details per each trial, this objective function can also be another suitable baseline for assessing our proposed optimizer and nonlinear objective function.

### 4.1.1.4. Weighted Common Spatial Pattern (Wgt-CSP), our proposed objective function as an optimizer evaluator

To evaluate the effectiveness of OEO in parameter learning schemes, the algorithm has to outperform non-hybrid GD, GD with PSO as initializer (GPSO) [35] and some other promising GD-hybrid methods. So, a differentiable objective function is used that has a multi-modal search space. The proposed objective function tries to tackle the challenge of improving EEG evaluation-data classification accuracy by more informative feature selection. By using a suitable initializer (i.e. a meta-heuristic) in GD, this challenge will be handled. This new objective function is a revision for CSP. Its purpose is to:

$$\frac{w^T (\sum_{i=1}^{n} a_i \Sigma_{i,c}) w}{w^T (\sum_{i=1}^{n} b_i (\Sigma_{i,1} + \Sigma_{i,2})) w} \tag{7}$$

Where $\Sigma_{i,c}$ is epoch's covariance matrix of class c as NxN matrix with N as the number of channels. w is one row of CSP projector. This process is to lower the total number of epochs either for the sake of increasing precision or for passing a smaller number of covariance matrices to CSP algorithm. Due to the usage of GD for optimization, negative of (7) will be minimized in terms of $\underline{w}$, $\bar{a}$ and $\bar{b}$.

Using Rayleigh Quotient, both of the following equations, when having the weights $a_i$ and $b_i$, can be simplified to a generalized eigenvalue problem and the projectors be optimized in a definite way. That makes the only uncertain part of optimization be the process of finding $a_i$ and $b_i$ s. So unlike Devlaminck et. al method which has the likelihood of getting caught in local optima in all its parameters, this form helps the process of optimization be done in a more robust way while remaining more certain about the result. A schematic of the proposed objective function is shown in Figure (4).

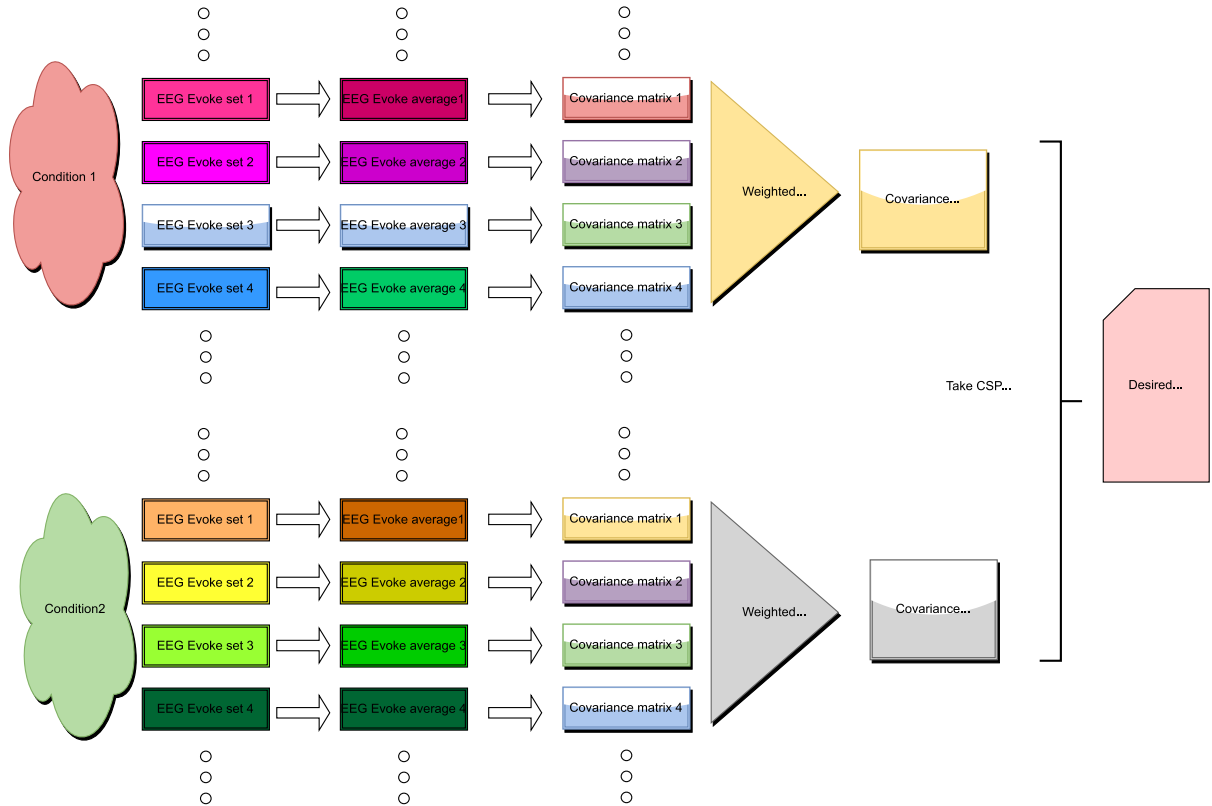

**Figure. (4). General schema of the proposed method.**

### 4.1.2. Optimization and evaluation pseudo-codes, problem-specific tuned-up hyper-parameters

All OEO hyper-parameters are set to the values from Table 3. The GD-hybrid algorithm is discussed in 5.1.2.1. The cross-validation procedure for validation accuracy assessment of data is explained in 4.1.2.2. Lastly, the GD parameter specification phase is shown in 5.1.2.3.

### 4.1.2.1. Hybrid optimizer pseudo-code

Due to non-convexity of Wgt-CSP objective function, mere usage of GD without powerful random initializer cannot lead to global minima and as a result, GD suffers a premature convergence.

The main optimization process alternates between computing CSP using SVD power method with 7 iterations and OEO-GD for updating weights with 6 iterations. The OEO-GD algorithm is mentioned as below:

**Algorithm 2, The OEO-GD algorithm**

**Input:** $\bar{a}$, $\bar{b}$ : weights in (7) . N: number of GD iterations as 6. F: Objective function in (7). $\gamma$ : 1e-10 . Parameter settings for OEO.

**Output:** Optimized $\bar{a}$, $\bar{b}$, W

**Pseudo-Code:**
1. Normalize and remove means of mixed components
2. Randomly pull a matrix with positive elements.
3. Randomize $\bar{a}$, $\bar{b}$ by uniformly randomized initialization.
4. Do until $iter < Max\_Iterations$ or $\sum_{i,j} abs(\bar{a}_{iter} - \bar{a}_{iter-1}) ) < \gamma$
    1. Initialize $\bar{a}$, $\bar{b}$ by OEO algorithm.
    2. Do until i < N
        1. Compute $\nabla_W F$ using computed gradient.
        2. Remove corrupted (high values / NaN values) columns of $\nabla_W F$
        3. Set $\underline{a}_{iter} = \underline{a}_{iter-1} - \eta * \nabla_W F$
           $\underline{b}_{iter} = \underline{b}_{iter-1} - \eta * \nabla_W F$
        4. Compute new CSP projector using eigenvalue decomposition.
        5. set i = i+1

---

**Algorithm 3, the 5-folded cross-validation process for classifying EEG signals**

**Input**: Data, Labels, and Classifiers
**Output**: Averaged Classification Accuracy

**Pseudo-Code**:
1. Randomly divide trials into 5 folds of trials with the same length.
2. Normalize and remove mean of each data channel.
3. For each selected fold; do:
    1. Use the selected fold as the evaluation set and others as train data.
    2. Decimate data by 8. Bandpass filter the data using a fifth order Chebyshev filter with band-pass of [3, 30] Hz.
    3. Learn a CSP projector using training data folds using CSP algorithm described in 5.1.1.1.
    4. Extract significant components of data out of 16 from both train and test data.
    5. Pass the results in train and test data to Linear Discriminant Analysis (LDA) or Support Vector Machine (SVM) classifier with linear kernel and LibSVM library [43].
    6. Save the validation accuracy
    4. Average over 5 resulted accuracies.

.

### 4.1.2.2. EEG dataset and evaluation approach used for the proposed CSP objective function

EEG data is derived from a speech-imagery BCI experiment developed by Rostami et.al. [40]. The Data is recorded by a 16 channeled EEG meter extracted from 6 subjects, aged between 23 and 30 who performed imagination of vowel sounds. Each subject has taken 180 trials which were approximately 36 trials for the imagination of five class each as a vowel. In this paper, only two class is used for classification which was classes 2 and 5. The sampling rate was 512 Hz for 4 seconds lasted imagery. The main 5-folded cross-validation process is described as follows:

### 4.1.2.3. Values used for searching and tuning hyper-parameter

For GD, Adam is used due to tuning outperformance versus other gradient methods. The tuner has examined learning-rates {0.2,0.02,0.002}, checked {0.8,0.9,099} for momentum1 and momentum2, and the proposed OEO initializers' recurrence occasion are fine-tuned as 3, 6, 9, 12 and 15 GD runs. The best-resulted parameters settings are 0.2 for learning-rate, 0.9 for momentum1, 0.9 for momentum2 and 6 for initializer rerun occasion.

The error-bar in Figure 5 shows that the proposed hybrid optimizer of Wgt- CSP over sections has made it outperform other similar averaging methods in 5.1.1.2 and 5.1.1.3. All accuracies in the error-bar are validation data averaged over 20 independent runs in the first three subjects of speech imagery dataset [40]. All methods except Wgt- CSP are evaluated by sole GD without any meta-heuristic initializers due to their convex structure. Eigen-decomposition power method was the CSP solver in the all aforementioned objective functions.

### 4.1.3. Results and comparisons

Table 4 shows elapsed time, cost and standard-deviation averaged over 20 independent runs and also best accuracy evaluated on validation data. Two best results per each column are shown in bold. The comparison baseline for the proposed objective function is the single GD with uniform random initialization, GPSO, and ICA-GD with GD as their main algorithm [36, 35, 44]. GPSO [35] is a combination of GD optimizer with standard Particle Swarm Optimization (PSO), which lacks multi-scaled search capability. Selection of PSO as baseline is due to its frequent usage in the contexts of hybrid engineering-based optimizations [35]. ICA-GD is a combination of Imperialistic Competitive Algorithm (ICA) [44] with GD as its main algorithm. ICA is used for comparison due to its simplicity and easy implementation in neural network and machine learning scenarios [52]. In GPSO, initialization frequency is controlled by the parameter $N_G$ [35]. Hence this parameter is set to 6. Results show outperformance of both OEO-GD and M-OEO-GD versus GPSO and single GD. Moreover, the generalization accuracy of classification is increased when OEO is added as a random initializer to GD. In addition, by using ICA as a GD initializer with the same routine as in OEO-GD, OEO is still the winner compared to ICA.

The validation accuracies also suggest that the reduction in the cost of the objective function over hybrid-GD mode is meaningful and that improves the best GD evaluation accuracy mentioned in $3^{th}$ row of Table 4.

**Table 4. Comparison of hybrid optimizers in terms of elapsed time, cost, and accuracy of validation data.**

| Optimizer method | Averaged elps time | cost mean (Proposed Wgt-CSP objective function. (Minus of Obj-fun in Formula (7)).) | cost StD. | Best Eval. Accuracy |
|---|---|---|---|---|
| OEO-GD* (Proposed) | 0.49 | **-0.896** | 0.04 | **86.12** |
| M-OEO-GD* (Proposed) | 0.53 | **-0.897** | 0.04 | **86.95** |
| GD | **0.49** | -0.876 | 0.05 | 79.60 |
| ICA-GD (ICA: [44]) | 0.83 | -0.863 | **0.04** | 75.81 |
| GPSO [34] | 0.59 | -0.892 | **0.04** | 81.70 |

Furthermore, Fig. 5 describes the Comparison of generalization accuracy in our Wgt-CSP method among previous subspace filtering approaches in EEG two-class classification study. Wgt-CSP outperformed other methods mostly when we used simple OEO-GD and M-OEO-GD as the optimizer.

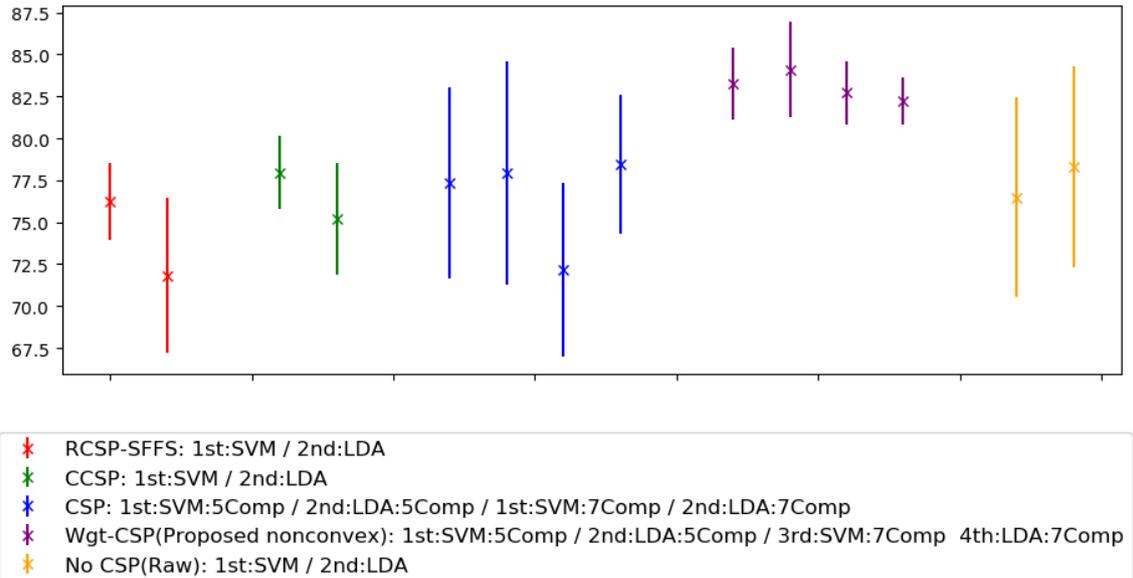

**Figure (5).** Comparison of EEG classification accuracy for weighted-CSPs-over-trials (Wgt-CSP) objective function among previous subspace approaches. RCSP-SFFS is Guan's work in 5.1.1.3. CCSP objective function of Section 4.1.1.2.Wgt-CSP while optimized by M-OEO-GD, outperformed other methods. In the legend, terms 'Comp', 'SVM', 'LDA', 'Raw' mean 'number of projected subspaces', 'Support Vector Machine', 'Linear Discriminant Analysis', and 'feature extraction without CSP' respectively.

### 4.2. Parameter Tuning Study, DGs Integration

Rather than approximative gradient-based problems with nonhomogeneous peaks distributed in a region, there are large scale engineering problems that are non-exact, nonanalytic while containing oddly distributed peak densities. For such problems, due to lack of convexity in sub-regions, one cannot use GD to speed up the process. However, the use of OEO in such problems may hinder from wasting time in the small region due to its useful localized memory control mechanism.

#### 4.2.1. Problem statement

A large amount of generated power in power stations is lost in the transmission systems. The losses not only waste electrical energies but also, occupy transmission line capacity during energy transmission. DGs are able to decrease power losses by placing optimally in the right situations in power systems. The large-scale calculation must be performed in distribution power system analysis. Therefore, an optimization method with an acceptable convergence rate can decrease the power system calculation time. OEO is used to decrease the cost function more than commonly used optimizers that are reasonably faster.

The recent developments in the electrical energy generation methods have brought about a new arena in power systems analysis, which is aptly called the distributed generation studies. Actually, DGs not only have a lower impact on the environment and produce clean energy, but also they are able to reduce the power system losses and increase the system reliability. An inappropriate placement of a DG can cause a wide range of harmful effects in the power network instead of benefits the system. For this reason, engineers have to determine the best sizes and locations for DGs to improve the power quality and system reliability. On the other hand, electrical power networks are created based on complex computation processes. A power system analyzer, therefore, in addition to powerful computers requires fast and high precision software and tools to analyze the network in a possible minimum time.

In conventional power systems in absence of any kinds of DG, the power buses in the system substations have a linear behavior in terms of the distance between the power plants and the investigated bus or buses. In other words, the increment in distance between the generator and the load leads to a decrease in the number of bus voltages in the load location. Thus, it seems logical to analyze the network by an algorithm which is able to use line search during its optimization process for seeking the optimum solutions. Moreover, there may be a wide range of buses near the power plants in a power system. In this condition, DG construction on these types of buses can cause overvoltage problem which decreases the power system stability and its reliability. Hence, an optimization method with an ability to search in a vast area of a search space in a little time can increase the power system analysis pace to assess the DGs places and sizes. The OEO, accordingly, can be a qualified choice as a power system analyzer tool because of its ability to seek the optimum answer in minimum time by employing a special useful memory in a wide search space. Therefore, it is expected that the OEO algorithm works as a fast and exact optimizer to find the best solutions for the places and sizes of DGs.

The proposed OEOs are applied to the 69-IEEE power distribution system to determine the optimum locations and sizes of the DGs to reduce the power system losses and improve voltage in four different scenarios. It is done in order to decrease power system losses profile during operation. This section demonstrates the load flow calculation process to derive the final loss function. Section 4.2.2. discusses the optimization results to demonstrate the effectiveness and applicability of the OEO in power system analysis usage and DGs sizing and placing in a power network.

Nowadays, DGs integration is increased in power distribution systems to improve power quality and voltage stability of a grid. DG costs and volumes are impressively lower than high capacity power stations. For this reason, they have found Special popularity among governments and users. DGs can bring a wide range of facilities for a power system. For instance, solving economic problems for generation development, environmental pollution reduction, improvement of power quality for users, power system losses reduction, voltage profile improvement, and recovering power system capacity are all privileges of a DG connected to a grid. Although DG foundation entails lots of facilities for a grid, engineers must determine a suitable place for the DG establishment in the grid. If a DG is erected at a wrong place in a grid, it is not only will not be useful, but also, it can bring up some problems in the system such as overvoltage and instability problems. On the other hand, power system analysis is of great complexity because of its iterative calculation process. Thus, engineers must spend a lot of time to find an optimum place for a DG in a grid analytically. Therefore, analyzing highly sophisticated power distribution systems with a wide range of electrical elements such as different loads, generators, and various DGs have a great difficulty and engineers need powerful computers to solve these kinds of problems. Hence, a useful optimization method can speed up power system analysis in the presence of nonlinear and different power system elements. There is a large number of pieces of literature considered various conditions for a power system to locate DGs in grids by means of meta-heuristic algorithms (e. g. [25-29]).

#### 4.2.1.1. Power System Analysis and system test
The 11kV, 69 buses radial electrical power distribution system [30, 33] is chosen as a case study network in this paper. The fast decoupled load flow method which has been proposed in [31] is used to solve the load flow problem. According to Figure 6, DGs inject electrical power in each bus during the optimization process. The injected active and reactive powers of the DGs are formulated by (8).

$$S_s = P_s + jQ_s = (P_L - P_{DG}) + j(Q_L - Q_{DG}) \tag{8}$$

Where $S_s$ is the power system apparent power which flows to the selected bus, and $P_s$ and $Q_s$ are the power system active and reactive powers, respectively. The load active and reactive powers are shown by $P_L$ and $Q_L$, respectively. Also, $P_{DG}$ and $Q_{DG}$ are the DG generated active and reactive powers in each selected bus during optimization. The lost power in the power system analysis can be calculated by (9).

$$P_{LOSS} = \sum_{i=1}^{n} P_{Gi} - \sum_{i=1}^{n} P_{Di} \tag{9}$$

Where $P_{Gi}$ and $P_{Di}$ are the generated and demanded powers on the ith bus respectively. Also, n represents the number of buses in the distribution network. On the other hand, there are some constraints which must be considered in the power system analysis. Active and reactive generated powers constraints for each substation in the network, transmission line capacity limit, and overvoltage consideration are four important limitations with a certain minimum and maximum values in the load flow calculation process.

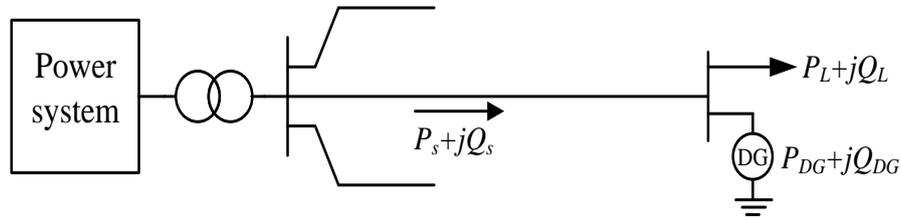

**Figure 6. A DG system connected to a grid bus**

The objective function of the optimization problem is shown in (10), which consists of DGs cost, penalty factor, and the network buses voltages.

$$OF = k_1 \times \sum_{i=1}^{n_{bus}} |V_i - 1| + K_2 \times S_{loss} + \sum_{i=1}^{n_{DG}} C_{DG}^i + PF \tag{10}$$

Where $V_i$ is the buses voltages in per-unit and $S_{loss}$ is the total losses in the network. Also, $n_{bus}$ and $n_{DG}$ represent the number of power system buses and the number of added DGs to the network, respectively. Moreover, C is the cost of each DG inserted to the network. $K_1 = 1.4e5$ and $K_2 = 1.35e5$ are the weighting coefficient for each term. Besides, PF shows the voltage violation penalty factor in each bus which must be formulated by (11).

$$PF = \sum_{i=1}^{n_{bus}} 1e6 \times [(V_i > 1.05) \| (V_i < 0.95)] \tag{11}$$

**4.2.1.2. Parameters to optimize**

In this study, the first and modified OEO proposed versions are tested to find the suitable size and place of the DGs in the radial power distribution network. In addition, the output of proposed optimization methods is compared with the particle swarm optimization (PSO) and Bat optimization approaches in the next section. The following instruction is executed for each optimization algorithm to find the best DGs location and sizes.

- Determine the number of DGs in the system
- Determine the minimum and maximum size of DGs
- Provide the power system information such as PV and PQ and slack buses, transmission line parameters, and the amount of generation and consumption on each bus
- Initialize the variable parameter such as DGs size and locations
- Apply load flow on the new system

- Consider the optimization problem cost function
- Apply optimization algorithm to find the best size and location of DGs

Also, All OEO hyper-parameters are set to the values from Table 3.

**4.2.2. Results and discussion:**

The power system analysis is performed in four different scenarios on the 69-IEEE test system. In each scenario, the number of DGs added to the grid is different from other scenarios. The power system losses, DGs cost, and capacity, and the minimum and maximum values of the voltage profile are considered in the optimization process by applying (10) and (11). The determination of the best value of *G, B,* and *A* in OEO algorithms are important to achieve the minimum cost during the optimization process. For this reason, the tuning procedure to assign the suitable values of these parameters is run on the power system at the first step.

OEO and baselines, i.e. the conventional PSO and Bat algorithms, have solved the power system problem to find suitable places and sizes for DGs in the same iterations. As the baseline optimizers are favored by researchers due to lower runtime and performance on power system analysis applications, they are selected for evaluation comparison.

According to table 3, which shows the best value of the parameters in each optimization scenarios, there is a little difference between the values. Therefore, they can be considered as static factors during a complex optimization problem. The amount of G, B, and A parameters in the power system analysis stage are equal to 0.05, 0.01, and 0.5450 for both OEOs, respectively. The optimization process is executed in 45 iterations for each algorithm. Table 4 presents the important outputs, consisting of the final cost values, selected buses, and the chosen DGs capacity on each bus in all scenarios, after solving the optimization problem and final load flow process. The presented data in Table 5 corroborate the preference of the M-OEO in seeking the minimum cost in comparison to other algorithms.

**Table 5. The best values for tuned parameters in four scenarios**

| Scenario | Algorithm | G | B | A |
|---|---|---|---|---|
| Scenario NO. 1 | OEO | 0.29 | 0.01 | 0.5450 |
| | M-OEO | 0.05 | 0.01 | 0.5450 |
| Scenario NO. 2 | OEO | 0.05 | 0.01 | 0.7250 |
| | M-OEO | 0.05 | 0.01 | 0.5450 |
| Scenario NO. 3 | OEO | 0.05 | 0.01 | 0.5450 |
| | M-OEO | 0.05 | 0.01 | 0.5450 |
| Scenario NO. 4 | OEO | 0.05 | 0.01 | 0.3650 |
| | M-OEO | 0.05 | 0.01 | 0.0050 |

Furthermore, M-OEO superiority can be proven by comparing the performance of the proposed methods with three other commonly used optimization methods for the DG assessment in the power system. Pursuant to Table 6 which shows the results of load flow in presence of the determined DGs in each scenario, the proposed options defined by the M-OEO not only reduce the total power system losses but also improves the voltage profile in the system. The minimum and maximum voltage parameters in Table 7 show the amount of lowest and highest voltages on the power system after DGs placement. These parameters are equal to 0.92202 and 1 per unit for the power system in the absence of any DG. Furthermore, the total power system losses in the standard network (without DG) is 214.3878 Mw. Little differences between the minimum and maximum voltage magnitudes in each power system load flow and the voltage profile curves presented in Figures 7 to 10 demonstrate the modified OEO ability in finding best solutions to improve the voltage profile. Although in some scenarios in Table 6 the M-OEO power losses are bigger than PSO, the total costs of the M-OEO are dramatically lower in that scenario according to Table 6. It must be remembered that the network loss consideration is one of the objective function terms. Thus, the smaller costs of the M-OEO in each scenario demonstrate its ability to place DGs optimally to achieve the best voltage profile while the network losses are acceptably low. It must be noticed that the one kilo-volt-ampere DG price is assumed to be equal to 700$ in the simulation process and DG cost calculation.

**Table 6. The optimum costs and solutions after 45 iterations for each algorithm**

|  |  | Selected Buses | DGs capacities (w) | Minimum costs |
|---|---|---|---|---|
| **Scenario NO. 1** | Simple OEO | 12 | $2.1415*10^3$ | $5.9885*10^6$ |
|  | **M-OEO** | 60 | $2.1946*10^3$ | $\mathbf{5.9879*10^6}$ |
|  | PSO | 62 | $2.1516*10^3$ | $6.2679*10^6$ |
|  | BAT | 12 | $1.7884*10^3$ | $6.3929*10^6$ |
| **Scenario NO. 2** | Simple OEO | 5, 14 | $1.9744*10^3$ | $9.3855*10^6$ |
|  | **M-OEO** | 10, 60 | $1.8252*10^3$ | $\mathbf{9.3386*10^6}$ |
|  | PSO | 10, 61 | $1.5357*10^3$ | $1.4912*10^7$ |
|  | BAT | 10, 62 | $2.0046*10^3$ | $1.1658*10^7$ |
| **Scenario NO. 3** | Simple OEO | 60, 62, 65 | $1.0181*10^3$ | $\mathbf{1.0384*10^7}$ |
|  | **M-OEO** | 12, 59, 60 | 985.1606 | $1.0672*10^7$ |
|  | PSO | 12, 58, 60 | $1.1257*10^3$ | $1.1803*10^7$ |
|  | BAT | 35, 62, 66 | $1.4583*10^3$ | $1.6024*10^7$ |
| **Scenario NO. 4** | Simple OEO | 12, 47, 59, 60 | 979.1621 | $1.4197*10^7$ |
|  | **M-OEO** | 12, 59, 60, 61 | 774.4883 | $\mathbf{1.2204*10^7}$ |
|  | PSO | 12, 52, 59, 62 | 805.5472 | $1.5312*10^7$ |
|  | BAT | 27, 28, 57, 60 | $1.4663*10^3$ | $2.5214*10^7$ |

**Table 7. Network parameters in various scenarios and different algorithms**

|  |  | Minimum Voltage among all buses (p.u.) | Maximum Voltage among all buses (p.u.) | Network losses (Mw) | DG prices ($) |
|---|---|---|---|---|---|
| **Scenario NO. 1** | Simple OEO | 0.93707 | 1.0066 | 155.1814 | 1499050 |
|  | M-OEO | 0.9721 | 1 | 45.0773 | 1536220 |
|  | PSO | 0.97188 | 1 | **34.6167** | 1506120 |
|  | BAT | 0.93467 | 1.0005 | 153.5686 | **1251880** |
| **Scenario NO. 2** | Simple OEO | 0.93627 | 1.0242 | 181.0494 | 2764160 |
|  | M-OEO | 0.98844 | 1.0071 | 37.949 | 2555280 |
|  | PSO | 0.98437 | 1.0019 | **26.8368** | 2149980 |
|  | BAT | 0.99428 | 1.0104 | 30.1606 | 2806440 |
| **Scenario NO. 3** | Simple OEO | 0.97771 | 1.0261 | 50.4389 | 2138010 |
|  | M-OEO | 0.9845 | 1 | **32.4693** | **2068837** |
|  | PSO | 0.99054 | 1.0041 | 33.9089 | 2363970 |
|  | BAT | 0.98239 | 1.044 | 91.1834 | 3062430 |
| **Scenario NO. 4** | Simple OEO | 0.98417 | 1 | 32.4856 | 2741654 |
|  | M-OEO | 0.98701 | 1.0037 | **22.7675** | 2168567 |
|  | PSO | 0.98243 | 1.0017 | 29.0284 | 2255532 |
|  | BAT | 0.99431 | 1.05 | 98.0849 | 4105640 |

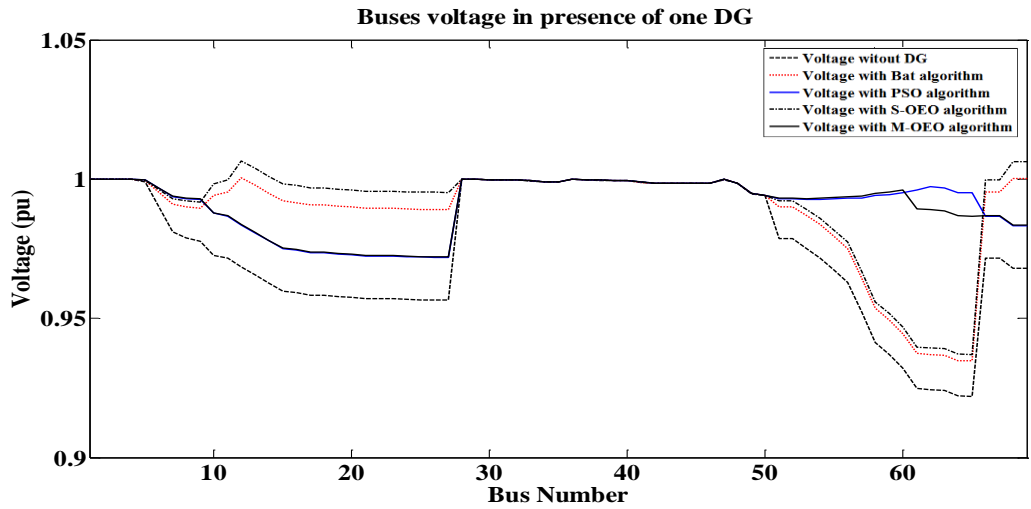

**Figure 7. Voltage profiles in presence of one DG after determination of the optimum solutions with each optimizer**

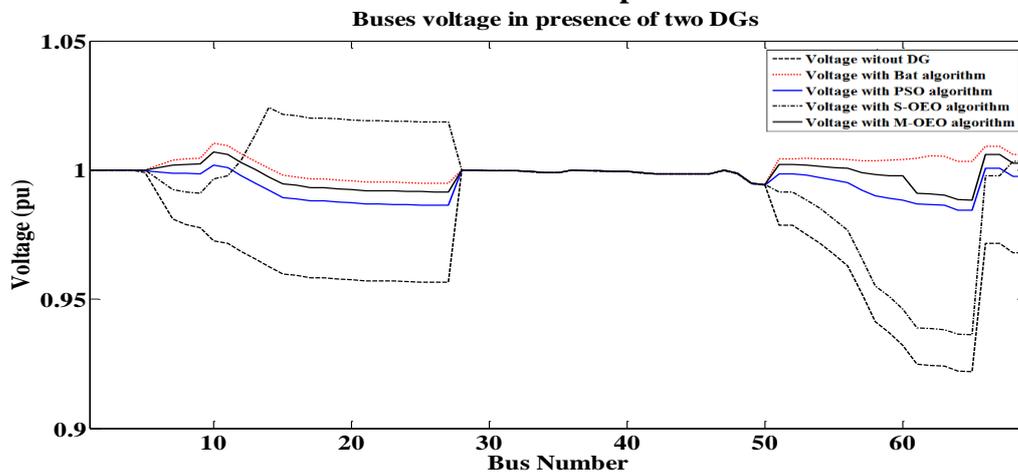

**Figure 8. Voltage profiles in presence of two DGs after determination of the optimum solutions with each optimizer**

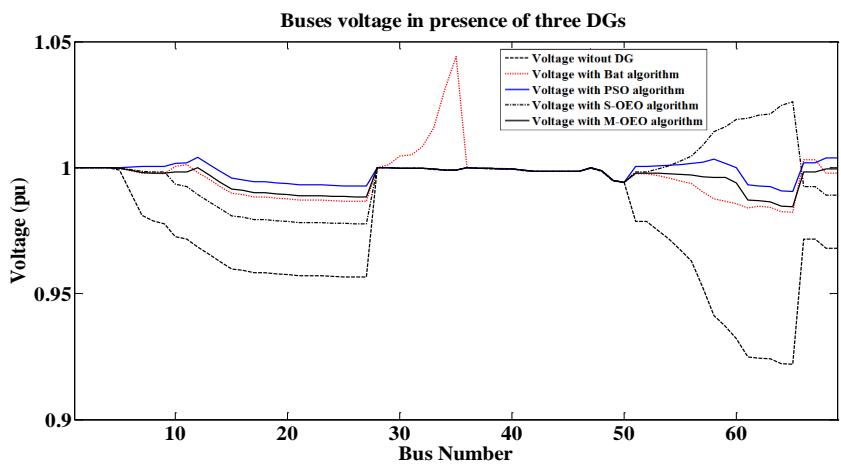

**Figure 9. Voltage profiles in the presence of three DGs after determination of the optimum solutions with each optimizer**

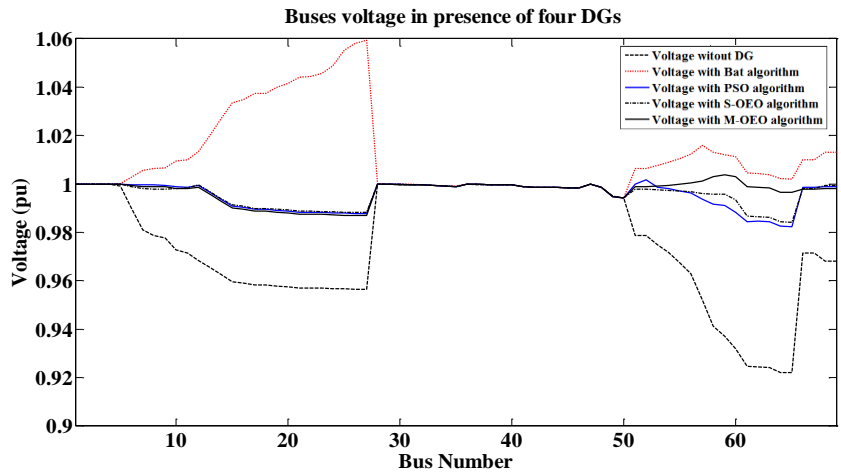

Figure 10. Voltage profiles in the presence of four DGs after determination of the optimum solutions with each optimizer

## 5. Conclusion:

One of the most important issues which must be addressed in meta-heuristic methods is an adaptive controlling extent of memory entering the solution decision during optimization. Unfortunately, this problem has not been issued wholly. The OEO is an attempt to harness the problem by inspiring from a concept in modern physics called observer effect. The new algorithm has been checked out in term of effectiveness with two types of real-world applications, hybrid with GD mode and single tuning mode. The results are fairly acceptable and show that the OEO can outperform algorithms with disorderly distributed peaks and leads to informative spaces while seeking solutions in such complex cost functions. The proposed algorithms worked more effectively in 3 benchmark functions with non-homogeneous peaks distributions comparing to 2 benchmark functions with homogeneous peaks.

For the first real-world scenario as a hybrid optimizer of EEG features classification, OEO has overcome GD and two basic simple hybrids with GD. That promising result with small relative elapsed time will be motivation for combining metaheuristics with gradient-based algorithms in a larger amount of problems. Due to the formulation of CSP subspace filtering problem, the GD version used in the proposed study was not stochastic like ones used in deep learning; but averaging over sub-batches and saving average cost per batch, may also help big-data models initialization. Such large scale models will be optimized by OEO-GD in our future work.

For the second real-world scenario as parameter tuner, OEO and baselines are used. According to the results, the OEO has found better solutions for the system. Therefore, the OEO is able to solve complex problems in comparison to the PSO. Thus, scientists and engineers can spend less time to find suitable solutions for complex and sophisticated mathematics problems.

In the future work, the role of the Modified OEO can be to inject a constraint in the adaptation of G(i) in i'th cluster such that all values for neighboring clusters remain close to each other. This may result in more reasonable solution update and exerting more regionally smooth observer effect. In the upcoming works, a more comprehensive parameter tuning will be approached and new cases of nonlinearity will be proposed. Also, tuning of nonlinearity parameters will be set as criteria for evaluation of nonlinearity.

**Conflict of Interest:**

The authors declare that they have no conflict of interest.